\documentclass{article}
\usepackage{spconf,amsmath,graphicx,hyperref}
\usepackage{multirow}
\usepackage{booktabs}
\usepackage{verbatim}
\usepackage{xcolor}

\title{ECG-Agent: On-Device Tool-Calling Agent for ECG Multi-Turn Dialogue}
\name{\shortstack{Hyunseung Chung$^{1}$, Jungwoo Oh$^{1}$, Daeun Kyung$^{1}$, Jiho Kim$^{1}$, \\
                 Yeonsu Kwon$^{1}$, Min-Gyu Kim$^{2}$, Edward Choi$^{1}$}}

\address{%
  \centerline{$^{1}$KAIST \quad $^{2}$Ajou University School of Medicine}
}
\begin{document}
\ninept
\maketitle
\begin{abstract}
Recent advances in Multimodal Large Language Models have rapidly expanded to electrocardiograms, focusing on classification, report generation, and single-turn QA tasks. However, these models fall short in real-world scenarios, lacking multi-turn conversational ability, on-device efficiency, and precise understanding of ECG measurements such as the PQRST intervals. To address these limitations, we introduce ECG-Agent, the first LLM-based tool-calling agent for multi-turn ECG dialogue. To facilitate its development and evaluation, we also present ECG-Multi-Turn-Dialogue (ECG-MTD) dataset, a collection of realistic user-assistant multi-turn dialogues for diverse ECG lead configurations. We develop ECG-Agents in various sizes, from on-device capable to larger agents. Experimental results show that ECG-Agents outperform baseline ECG-LLMs in response accuracy. Furthermore, on-device agents achieve comparable performance to larger agents in various evaluations that assess response accuracy, tool-calling ability, and hallucinations, demonstrating their viability for real-world applications. The dataset and code are available in \href{https://github.com/gustmd0121/ECG-Agent}{this URL}
\end{abstract}

\section{Introduction}
\label{sec:intro}

In recent years, the demand for on-device language models has increased exponentially as users shift toward personalized experiences on smartphones and wearables, moving away from traditional Large Language Models (LLMs) or Multimodal Large Language Models (MLLMs) \cite{xu2024device}. On-device LLMs offer superior privacy, faster response speed, and offline accessibility \cite{zheng2025review}, making them especially valuable in personal health monitoring. A key application is ECG interpretation via wearable devices, where on-device LLMs can transform cardiac monitoring into personalized health insights \cite{wang2024efficient} and provide user-friendly explanations \cite{nissen2025medicine}. 
Despite the clear benefits of on-device LLMs for ECG interpretation, most recent advancements have come from ECG-LLMs largely inspired by the Large Language and Vision Assistant \cite{liu2023visual} architecture. These models have demonstrated strong performance in various tasks such as classification, report generation, and single-turn QA \cite{zhao2024ecg, liu2024teach, lan2025gem, yang2025ecg}. However, they struggle to meet requirements for practical use. 

First, natural user-assistant communication is inherently multi-turn, yet existing models are limited by single-turn interaction, as they are trained on isolated examples and fail to consider prior context. For example, PULSE-7B, trained on the single-turn ECGInstruct dataset \cite{liu2024teach}, performs poorly on the multi-turn dialogue benchmark we constructed in this work, where each follow-up question depends on prior turns. Second, on-device deployment requires lightweight models with low memory and compute demands, but current architectures, coupling a pre-trained ECG encoder to a large LLM (often $>$7B parameters), are too large for resource-constrained devices. Standard smartphone RAM (6–8GB) is the practical bottleneck: a 4-bit quantized 3B model requires approximately 2GB and fits comfortably alongside the operating system, whereas an 8B model demands 5–6GB, exceeding usable memory. Finally, user inquiries require precise, measurement-based responses, but these models suffer from lack of precision, as the ECG encoding process often sacrifices the fine-grained ECG details needed for measurements like PQRST intervals.

\begin{figure}[!t]
  \centering
  \includegraphics[width=\linewidth]{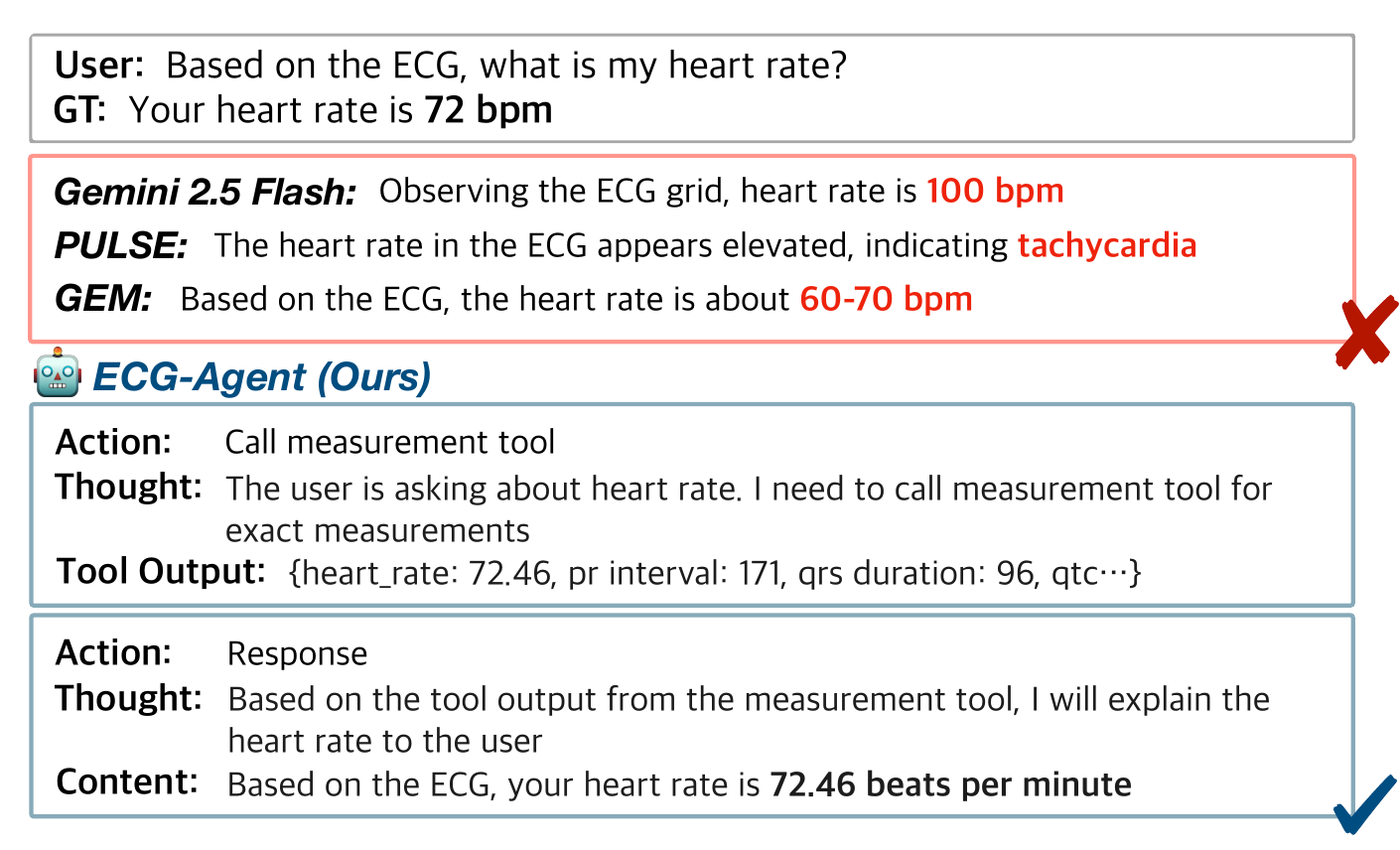}

  \caption{An example of a single conversational turn from a multi-turn dialogue in the ECG-MTD test set. The baselines output inaccurate responses to a user inquiry about heart rate, while the ECG-Agent (1B) produces a precise, measurement tool-based response.}
  \label{motivational}
\end{figure}

In contrast, LLM-based tool-calling agents offer a promising alternative. ``Tool-calling" leverages a core LLM for reasoning and planning, which in turn delegates specific tasks to external, specialized tools. This approach allows a unimodal, text-only LLM to effectively analyze complex, non-textual data by offloading tasks like signal interval measurement to a dedicated tool. By doing so, it not only enables the use of smaller backbone models of 3B parameter size or less for comparative performance, but also directly alleviates the lack of precision, as these tools can provide quantitative measurements to best answer user queries.  LLM-based tool-calling agents have proven effective for external API calls based on generated tool-calling datasets and evaluation benchmarks \cite{patil2024gorilla, qin2024toolllm, shimtooldial} in the general NLP domain. This trend expanded rapidly to the medical domain, and agents such as MMedAgent \cite{li2024mmedagent}, MedRAX \cite{fallahpourmedrax}, and FactCheXcker \cite{heiman2025factchexcker} demonstrate superior performance by integrating tools for clinical text and imaging modalities.

\begin{figure*}[!t]
    \centering
    \includegraphics[width=2.0\columnwidth]{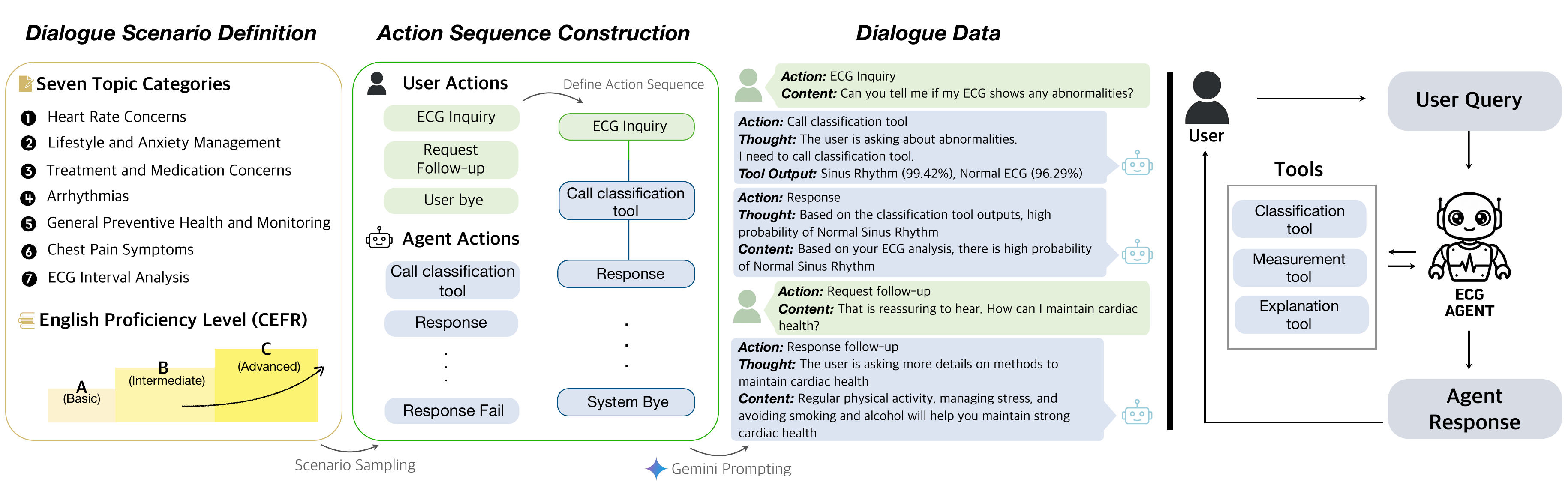}
    \caption{(\textit{Left}) Overall pipeline of ECG-MTD dataset construction. Dialogue Scenario defined by seven topic categories acquired from online medical consultation datasets and three English proficiency levels from the official CEFR guidelines. Action sequences constructed from pre-defined user and agent actions. A topic category, a user CEFR level, and an action sequence sampled to prompt Gemini-2.5-Flash to generate ECG multi-turn dialogue. \textit{(Right}) Training framework of ECG-Agent. User query is processed to the ECG-Agent, where the agent is able to use three different tools to generate a response to the user. This process is repeated in multiple turns during multi-turn dialogue.}
\label{model}
\end{figure*}

In this paper, we first introduce the ECG Multi-Turn Dialogue (ECG-MTD) dataset, designed to reflect realistic user-assistant interaction and equip agents with multi-turn conversational reasoning capabilities. Then, we propose ECG-Agent, which to the best of our knowledge is the first agent with tool-calling capabilities to address ECG multi-turn user queries. We conduct comprehensive evaluation of ECG-Agents across various parameter sizes (1B, 3B, 8B, 32B) against baseline ECG-LLMs, to show superiority of our agents and the comparable performance of on-device agents ($<$4B parameters). 

\section{Method}
\label{sec:method}

\subsection{ECG-MTD Dataset}
To train and evaluate a tool-calling agent, we constructed the ECG-MTD, a multi-turn dialogue dataset where an agent must select appropriate actions. We analyzed two real online medical consultation datasets HealthCareMagic-100k and icliniq-10k \cite{li2023chatdoctor}. After filtering for ECG-related keywords, we went through a detailed qualitative analysis of approximately 1000 representative conversations to determine representative topic categories. We then used the Gemini-2.5-Pro model \cite{comanici2025gemini} to refine our initial list of topics to finalize a schema of seven distinct and representative categories, as shown under Dialogue Scenario Definition (Figure \ref{model}). To ensure our dataset realistically captures a wide range of patient interactions, we diversified our generation process along two axes: conversational topic and user language proficiency. For language proficiency, we adopted the Common European Framework of Reference for Languages (CEFR), an international standard that we simplified into three levels: A (basic), B (intermediate), and C (advanced).

With the dialogue scenarios defined, the next step was to structure the agent's behavior. Since the agent must learn to perform tool-calling, we constructed action sequences that map out the flow of a conversation, dictating when the agent should call a specific tool or respond directly to the user. This process closely follows that of ToolDial \cite{shimtooldial}, for which we first defined three possible user actions (\texttt{ECG inquiry}, \texttt{request follow-up}, \texttt{user bye})  and seven agent actions. The agent actions include responding directly (\texttt{response}), failure to respond (\texttt{response fail}), responding to follow-up request without additional tool use (\texttt{response follow-up}), \texttt{system bye} to end dialogue and three tool actions (\texttt{classification, measurement, explanation}) shown in Figure \ref{model}. The \texttt{response fail} action is selected in the subsequent turn after the agent fails to produce valid tool outputs (e.g. null values).
From these actions, we compiled 20 realistic action sequences to model common multi-turn conversational flows, such as a user asking about the ECG, the agent calling the appropriate tool, and then responding with the result (e.g., $\text{ECG Inquiry} \rightarrow \text{Call Classification Tool} \rightarrow \text{Response} \rightarrow \text{User Bye} \rightarrow \text{System Bye}$).
  
Finally, we generated the ECG-MTD dataset by prompting the Gemini-2.5-Flash model \cite{comanici2025gemini}. The action sequence served as a skeleton by specifying the action for each conversation turn.
Any generated samples that deviated from the prompted format were subsequently filtered out.
To ensure a diverse dataset, each dialogue was generated from a unique combination of a topic category, a CEFR level, and an action sequence.
These components were sampled uniformly to define the dialogue's theme, the user's language proficiency, and the conversational structure, respectively. 
To control the user's language proficiency, the generation prompts included vocabularies corresponding to each CEFR level drawn from two sources: a standard CEFR-labeled dictionary for general vocabulary, and a custom list of 30 medical terms generated via GPT-4o for each CEFR level.
We generated dialogues based on the ECG samples from the PTB-XL dataset \cite{wagner2020ptb}.
To mirror both clinical and consumer wearable scenarios, we generated data for three distinct lead configurations: 12-lead (clinical standard), Lead I (wearables), and Lead II (patches).
To ensure medical accuracy for each scenario, we consulted a licensed physician to identify the specific set of FDA-cleared diagnostic classes detectable for each lead configuration from the 71 classes in the PTB-XL dataset.

\begin{table*}[!t]
\centering
\caption{ECG-MTD evaluation under 12-lead, Lead I, and Lead II configurations. All metrics are evaluated with Gemini-2.5-Pro. Accuracy and Completeness are scored against the generated ground-truth test set on a 1–5 scale. Reported averages for Accuracy and Completeness are computed over post classification and measurement tool responses.}
\label{tab:lead_results}
{\small
\setlength{\tabcolsep}{6pt}
\renewcommand{\arraystretch}{0.75}
\begin{tabular}{llccc|cccc}
\toprule
&& \multicolumn{3}{c|}{\textbf{Baselines}} & \multicolumn{4}{c}{\textbf{ECG-Agents (Ours)}} \\
\cmidrule(lr){3-5}\cmidrule(lr){6-9}
\textbf{Leads} & \textbf{Metric} & Gemini-2.5-Flash & PULSE & GEM & Llama 3.2 1B & Llama 3.2 3B & Llama 3.1 8B & Qwen3 32B \\
\midrule
\multirow[c]{2}{*}{12 Leads}
  & Accuracy       & 1.85    & 2.27    & 2.48    & 3.44    & 3.45    & \underline{3.50}    & \textbf{3.54}    \\
  & Completeness   & 2.26    & 1.87    & 2.02    & 2.59    & \underline{2.64}    & 2.57    & \textbf{2.65}    \\
\midrule
\multirow[c]{2}{*}{Lead I}
  & Accuracy       & 2.69    & 1.85    & 1.62    & \textbf{3.76}    & \underline{3.71}    & 3.51    & \textbf{3.76}    \\
  & Completeness   & \textbf{3.02}    & 1.61    & 1.47    & 2.70    & \underline{2.84}    & 2.75    & 2.80    \\
\midrule
\multirow[c]{2}{*}{Lead II}
  & Accuracy       & 2.52    & 1.88    & 1.64    & 3.76    & \textbf{3.88}    & 3.63    & \underline{3.81}    \\
  & Completeness   & \underline{2.95}    & 1.74    & 1.69    & 2.85    & \textbf{3.00}    & 2.84    & 2.94    \\
\bottomrule
\end{tabular}
}
\end{table*}
\vspace{-2mm}

\subsection{ECG-Agent}
There are three tools used by the ECG-Agent: A classification tool, measurement tool, and explanation tool. For the classification tool, we utilized the pre-training strategy from \cite{oh2022lead}, which is a self-supervised learning method that learns ECG representations by combining contrastive learning schemes and Random Lead Masking (RLM) to ensure robust classification performance in arbitrary lead configurations.
For the measurement tool, we used the Neurokit2 package \cite{Makowski2021neurokit}, which is a Python toolbox for neurophysiological signal processing. Finally, for the explanation tool, we utilized the Spectral eXplanation (SpectralX) framework \cite{chung2024time} which provides time-frequency explanations for time-series black-box classifiers. As SpectralX is designed for univariate time-series, the explanation tool was excluded for the multivariate 12-lead configuration. 

We instruction-tuned ECG-Agents in four different model sizes with the ECG-MTD dataset in 12-lead, Lead I, and Lead II configurations.
Using a turn-by-turn training method, we created training instances from each agent turn, with the full prior dialogue history as input.
Each user turn consists of an action and a content, and each agent turn consists of a triplet: an action, a thought (the agent's reasoning), and either a tool output or content (user and agent turns both shown in the Dialogue Data of Figure \ref{model}.)         
Through instruction-tuning on the ECG-MTD dataset, the ECG-Agent learns to generate a `thought' process that maps a user's query to a specific tool-call action or a direct response.
This enables it to decide whether to classify an arrhythmia using the pre-trained model, extract measurements with Neurokit2, generate an explanation via SpectralX, or respond directly.
After selecting a tool-call action, the agent learns to convert the tool's structured output into a natural language response.
During inference, this trained behavior allows the agent to dynamically reason about the conversational context and execute the appropriate tool call to address the user's query.

\begin{table}[t]
\centering
\caption{Evaluation of direct responses (\textit{i.e.}, non-tool-calling turns). Scores are averaged across 12-lead, Lead I, and Lead II configurations, on a 1-5 scale evaluated by Gemini-2.5-Pro.}
\label{tab:direct_responses}
{\small
\setlength{\tabcolsep}{6pt}
\renewcommand{\arraystretch}{0.70}
\begin{tabular}{lcc}
\toprule
\textbf{Model} & \textbf{Accuracy} & \textbf{Completeness} \\
\midrule
Gemini-2.5-Flash & \textbf{4.25} & \textbf{4.53} \\
\midrule
PULSE            & 1.86          & 1.74          \\
GEM              & 1.75          & 1.62          \\
\midrule
Llama 3.2 1B         & 3.71          & 2.57          \\
Llama 3.2 3B         & 3.68          & 2.57          \\
Llama 3.1 8B         & 3.59          & 2.58          \\
Qwen3 32B        & 3.78          & 2.60          \\
\bottomrule
\end{tabular}
}
\end{table}
\vspace{-2mm}

\section{Experiments}
\label{sec:experiments}
\subsection{Experimental Settings} 
We conduct experiments on our generated ECG-MTD dataset, which is derived from the PTB-XL. The dataset contains 21,837 multi-turn ECG dialogues, deconstructing each dialogue into independent training instances, resulting in 98k samples per lead configuration (12-lead, Lead I, and Lead II), and an average of 7.68 turns per dialogue. We split the datasets into training (80\%), validation (10\%), and test (10\%) sets. Experiments were conducted on RTX A6000s for Llama 3.2 1B, 3B, and 3.1 8B, and A100s for the Qwen3 32B model.

\begin{table}[t]
\centering
\caption{Per-class Temporal Intersection over Union (TIoU) (\%) of the explanation tool (SpectralX) for Lead I and Lead II.}
\label{tab:class_accuracy}
{\small
\setlength{\tabcolsep}{6pt} 
\renewcommand{\arraystretch}{0.70}
\begin{tabular}{lcc}
\toprule
\textbf{Diagnostic Class} & \textbf{Lead I} & \textbf{Lead II} \\
\midrule
Premature Atrial Contraction & 64.47 & 67.31 \\
Premature Ventricular Contraction & \textbf{74.12} & 72.25 \\
ST-segment Depression & 71.87 & \textbf{75.99} \\
\bottomrule
\end{tabular}
}
\end{table}
\vspace{-2mm}

\subsection{Training Setup}
For the base LLM models of the ECG-Agents, three Llama-based and one Qwen-3 model are used as follows: Llama-3.2-1B-Instruct (Llama 3.2 1B), Llama-3.2-3B-Instruct (Llama 3.2 3B), Llama-3.1-8B-Instruct (Llama 3.1 8B), and Qwen3 32B. We train all ECG-Agents with a batch size of 128.
For fine-tuning, LoRA \cite{hu2022lora} is applied with rank 16, $\alpha$=16, and AdamW 8-bit \cite{dettmers20228bit} optimizer, implemented using the Unsloth \cite{unsloth} framework for efficiency.
With 10 warm-up steps, training runs for 3 epochs with early stopping based on the validation loss evaluated every 1000 steps.
The experiment starts with a learning rate of $2 \times 10^{-4}$ and a weight decay of $0.01$, using a linear learning rate scheduler. We set the maximum sequence length to 4096 tokens to accommodate the dialogue history. 

\subsection{Evaluation}
We evaluate our ECG-Agent and the baselines in terms of response quality, next action prediction, faithfulness, and dialogue quality. 
For response quality, we measure \textbf{accuracy} and \textbf{completeness} using Gemini-2.5-Pro as an evaluator, based on the LLM-as-a-Judge framework \cite{zheng2023judging}. We assess three types of assistant turns: post-classification (responses following a classification tool call), post-measurement (responses following a measurement tool call), and direct responses (turns not requiring a tool). For baseline models, which do not use tools, we evaluate their direct answers to the same queries that trigger a tool call in our ECG-Agent. We generated ground-truth responses by prompting Gemini-2.5-Pro to answer each test set query using reference data: cardiologist-labeled diagnostic codes from PTB-XL for classification-based queries, and PQRST measurements from the Uni-G program in the PTB-XL+ dataset \cite{strodthoff2023ptb} for measurement-based questions. Accuracy represents how well the response matches the ground-truth, and completeness represents how the response covers key information in the ground-truth. The evaluator scores on a 5-point scale, where 1 represents the least score and 5 represents the best score. To validate the reliability of the LLM-as-a-Judge evaluator, we conducted a human evaluation on 300 randomly sampled dialogues (100 per lead configuration), where an annotator scored the accuracy and completeness of model responses relative to the ground truth on the same 5-point scale. We measure agreement between human judgments and the LLM-based evaluator using Spearman’s rank correlation \cite{liu2023g}. We observe a strong correlation for accuracy ($\rho=0.70$, $p<0.001$) and a moderate correlation for completeness ($\rho=0.46$, $p<0.001$), indicating that the automated evaluator serves as a reliable proxy for human judgment.
Additionally, we evaluate the accuracy of our explanation tool using Temporal Intersection over Union (TIoU) \cite{shou2016temporal}. This metric measures the overlap between the predicted and ground-truth time intervals for three representative diagnostic classes detectable in single-lead (Lead I and II) ECGs.

For \textbf{Next Action Prediction (NAP)}, at turn $k$, the model predicts the next action $\mathcal{A}_k$ given the dialogue history up to the previous turn $(\mathcal{H}_{k-1})$, the accumulated agent's reasoning steps up to the previous turn $(\mathcal{R}_{k-1})$, and the current user utterance $u_k$. We evaluate this in two settings from \cite{shimtooldial}: ``Without GT'', a real-world scenario and default setting where the agent conditions on its self-generated history ($\hat{\mathcal{H}}, \hat{\mathcal{R}}$), formalized as $\hat{\mathcal{A}}_{k}^{\text{w/o GT}} = \mathcal{M}(\hat{\mathcal{H}}_{k-1}, \hat{\mathcal{R}}_{k-1}, u_k)$; and ``With GT'', an upper-bound setting using the ground-truth history ($\mathcal{H}, \mathcal{R}$), formalized as $\hat{\mathcal{A}}_{k} = \mathcal{M}(\mathcal{H}_{k-1}, \mathcal{R}_{k-1}, u_k)$, where $\mathcal{M}$ is the agent's output. The reported NAP metric is the exact match accuracy between the predicted and the ground-truth actions from the test set.
\textbf{Faithfulness} measures whether the agent's response is accurately grounded in the preceding tool output, quantifying its tendency to hallucinate. Gemini-2.5-Pro evaluates this as a binary task, assigning a score of 1 for perfect alignment and 0 for any mismatch.
Dialogue quality assesses the overall conversational performance of the fully generated dialogue. Using Gemini-2.5-Pro as an evaluator, we score two criteria on a 5-point scale: \textbf{Naturalness}, for human-like conversational flow, and \textbf{CEFR Adherence}, for how well the model adapts to the user's language proficiency.

\begin{table}[!t]
\centering
\caption{Evaluation of Next Action Prediction (NAP) and Faithfulness. NAP reports the percentage of correct next-action predictions given dialogue history. Faithfulness reports the percentage alignment between tool outputs and the agent’s subsequent responses.}
\label{tab:nap_results}
\renewcommand{\arraystretch}{0.70}
{\small
\begin{tabular}{l l ccc}
\toprule
\multirow{2}{*}{\textbf{Leads}} & \multirow{2}{*}{\textbf{Model}} & \multicolumn{2}{c}{\textbf{NAP}} & \multirow{2}{*}{\textbf{Faithfulness}} \\
\cmidrule(lr){3-4}
 & & \textbf{w/o GT} & \textbf{w/ GT} & \\
\midrule
\multirow{5}{*}{12 Leads}
 & Gemini-2.5-Flash      & 70.84 & 71.66 & -- \\
 \cmidrule(lr){2-5} 
 & Llama 3.2 1B          & 93.94 & 95.93 & 85.29 \\
 & Llama 3.2 3B          & 94.33 & 95.41 & 88.12 \\
 & Llama 3.1 8B          & \textbf{94.95} & \textbf{97.23} & 87.66 \\
 & Qwen3 32B             & 94.23 & 95.77 & \textbf{88.98} \\
\midrule
\multirow{5}{*}{Lead I}
 & Gemini-2.5-Flash      & 67.42 & 66.72 & -- \\
 \cmidrule(lr){2-5} 
 & Llama 3.2 1B          & 91.79 & 92.72 & 91.12 \\
 & Llama 3.2 3B          & 91.73 & \textbf{94.19} & 91.97 \\
 & Llama 3.1 8B          & 91.81 & 93.36 & 91.27 \\
 & Qwen3 32B             & \textbf{93.03} & 94.07 & \textbf{93.04} \\
\midrule
\multirow{5}{*}{Lead II}
 & Gemini-2.5-Flash      & 68.41 & 69.55 & -- \\
 \cmidrule(lr){2-5} 
 & Llama 3.2 1B          & \textbf{92.16} & 93.07 & 91.56 \\
 & Llama 3.2 3B          & 88.53 & 90.63 & 92.35 \\
 & Llama 3.1 8B          & 90.07 & 92.70 & 92.52 \\
 & Qwen3 32B             & 91.13 & \textbf{93.43} & \textbf{94.31} \\
\bottomrule
\end{tabular}
}
\end{table}

\begin{table}[t]
\centering
\caption{Dialogue naturalness and CEFR level adherence on a 1-5 scale. Scores are averaged across 12-lead, Lead I, and Lead II configurations. Evaluated with Gemini-2.5-Pro.}
\label{tab:dialogue_quality}
{\small
\setlength{\tabcolsep}{6pt}
\renewcommand{\arraystretch}{0.70}
\begin{tabular}{lcc}
\toprule
\textbf{Model} & \textbf{Naturalness} & \textbf{CEFR} \\
\midrule
Gemini-2.5-Flash & 3.97 & \textbf{4.50} \\
\midrule
PULSE            & 3.67 & 3.57 \\
GEM              & 3.63 & 3.51 \\
\midrule
Llama 3.2 1B         & 3.95 & 4.32 \\
Llama 3.2 3B         & 3.96 & 4.37 \\
Llama 3.1 8B         & 3.95 & 4.38 \\
Qwen3 32B        & \textbf{3.98} & 4.41 \\
\bottomrule
\end{tabular}
}
\end{table}

\subsection{Experimental Results}
We compare ECG-Agent against a proprietary MLLM and two state-of-the-art ECG-LLMs. The proprietary MLLM is Gemini-2.5-Flash, a powerful general-purpose model without ECG fine-tuning. 
For ECG-LLMs, PULSE \cite{liu2024teach} is a 7B model trained on over one million ECG image–text pairs, and GEM \cite{lan2025gem} uses a 7B backbone with dual encoders for raw signals and images.      

First, we evaluate the response quality. The results in Table \ref{tab:lead_results} are the average over post-classification and post-measurement responses, and show that our ECG-Agents consistently outperform all baseline models in terms of \textbf{accuracy} and \textbf{completeness} across all three lead configurations except for completeness in Lead I. This indicates that the tool-calling approach yields more precise and comprehensive answers than end-to-end MLLM methods for user inquiries that require classification and measurement tools. For direct responses (Table \ref{tab:direct_responses}), Gemini-2.5-Flash performs best, but the ECG-Agents still outperform PULSE and GEM. Table \ref{tab:class_accuracy} displays the TIoU of the explanation tool in Lead I and Lead II, with strong performance up to 76\%.

Next, Table \ref{tab:nap_results} presents the \textbf{NAP} accuracy and \textbf{Faithfulness}.
All our instruction-tuned ECG-Agents achieve excellent scores (mostly $>$90\%) for the NAP accuracy, showing they have effectively learned to map conversational context to the appropriate action.
This stands in contrast to the general-purpose Gemini-2.5-Flash, whose lower accuracy underscores the importance of domain-specific fine-tuning for reliable tool use.
For Faithfulness scores, overall single-lead agents have better performance compared to 12-lead agents.
We hypothesize this is because the 12-lead configuration involves a significantly larger and more complex set of diagnostic classes, which increases the likelihood of hallucination.
For all lead configurations, Qwen3 32B agent achieves the best faithfulness performance.   

Finally, we evaluate the overall dialogue quality. In Table \ref{tab:dialogue_quality}, our ECG-Agents produce responses with high \textbf{Naturalness} scores, with the Qwen3 32B agent slightly outperforming the powerful Gemini-2.5-Flash baseline. While Gemini-2.5-Flash holds an edge in \textbf{CEFR Adherence}, likely due to its vast pre-training and extremely large parameter size, all our agents score relatively high, confirming their ability to adapt to the user's language proficiency. 
A key finding is that the on-device capable agents (1B, 3B) perform comparably to larger agents (8B, 32B), with minimal performance gaps across response quality, next action prediction, and dialogue quality. This demonstrates the effectiveness of using smaller models.

\section{Conclusion}
In this paper, we introduced ECG-Agent, the first tool-calling agent for multi-turn ECG dialogue, and the ECG-MTD dataset for its development. Our experiments demonstrate that by leveraging specialized tools, ECG-Agent generates conversational responses that are more accurate and comprehensive than those from baseline ECG-LLM models. Critically, our compact on-device agents (1B, 3B) achieved performance comparable to much larger models, establishing the practicality of using powerful yet efficient agents.

\section{Acknowledgement}
\label{sec:acknowl}
This work was supported by the Institute of Information \& Communications Technology Planning \& Evaluation (IITP) grant (No.RS-2019-II190075, No.RS-2022-II220984), National Research Foundation of Korea (NRF) grant (NRF-2020H1D3A2A03100945), and the Korea Health Industry Development Institute (KHIDI) grant (No.HR21C0198), funded by the Korea government (MSIT, MOHW).

\bibliographystyle{IEEEbib}
\bibliography{icassp_2026_refs,refs}

\begin{thebibliography}{10}

\bibitem{xu2024device}
Jiajun Xu, Zhiyuan Li, Wei Chen, Qun Wang, Xin Gao, Qi~Cai, and Ziyuan Ling,
\newblock ``{On-device language models: A comprehensive review},''
\newblock {\em arXiv:2409.00088}, 2024.

\bibitem{zheng2025review}
Yue Zheng, Yuhao Chen, Bin Qian, Xiufang Shi, Yuanchao Shu, and Jiming Chen,
\newblock ``{A review on edge large language models: Design, execution, and applications},''
\newblock {\em ACM Computing Surveys}, vol. 57, no. 8, pp. 1--35, 2025.

\bibitem{wang2024efficient}
Xin Wang, Ting Dang, Vassilis Kostakos, and Hong Jia,
\newblock ``{Efficient and personalized mobile health event prediction via small language models},''
\newblock in {\em {MobiCom}}, 2024.

\bibitem{nissen2025medicine}
Leon Nissen, Philipp Zagar, Vishnu Ravi, Aydin Zahedivash, Lara~Marie Reimer, Stephan Jonas, Oliver Aalami, and Paul Schmiedmayer,
\newblock ``{Medicine on the edge: Comparative performance analysis of on-device LLMs for clinical reasoning},''
\newblock {\em arXiv:2502.08954}, 2025.

\bibitem{liu2023visual}
Haotian Liu, Chunyuan Li, Qingyang Wu, and Yong~Jae Lee,
\newblock ``{Visual instruction tuning},''
\newblock {\em NeurIPS}, vol. 36, 2023.

\bibitem{zhao2024ecg}
Yubao Zhao, Jiaju Kang, Tian Zhang, Puyu Han, and Tong Chen,
\newblock ``{Ecg-chat: A large ecg-language model for cardiac disease diagnosis},''
\newblock {\em arXiv:2408.08849}, 2024.

\bibitem{liu2024teach}
Ruoqi Liu, Yuelin Bai, Xiang Yue, and Ping Zhang,
\newblock ``{Teach multimodal llms to comprehend electrocardiographic images},''
\newblock {\em arXiv:2410.19008}, 2024.

\bibitem{lan2025gem}
Xiang Lan, Feng Wu, Kai He, Qinghao Zhao, Shenda Hong, and Mengling Feng,
\newblock ``{Gem: Empowering mllm for grounded ecg understanding with time series and images},''
\newblock {\em arXiv:2503.06073}, 2025.

\bibitem{yang2025ecg}
Kai Yang, Massimo Hong, Jiahuan Zhang, Yizhen Luo, Suyuan Zhao, Ou~Zhang, Xiaomao Yu, Jiawen Zhou, Liuqing Yang, Ping Zhang, et~al.,
\newblock ``{ECG-LM: Understanding Electrocardiogram with a Large Language Model},''
\newblock {\em Health Data Science}, vol. 5, pp. 0221, 2025.

\bibitem{patil2024gorilla}
Shishir~G Patil, Tianjun Zhang, Xin Wang, and Joseph~E Gonzalez,
\newblock ``{Gorilla: Large language model connected with massive apis},''
\newblock {\em NeurIPS}, vol. 37, 2024.

\bibitem{qin2024toolllm}
Yujia Qin, Shihao Liang, Yining Ye, Kunlun Zhu, Lan Yan, Yaxi Lu, Yankai Lin, Xin Cong, Xiangru Tang, Bill Qian, et~al.,
\newblock ``{ToolLLM: Facilitating Large Language Models to Master 16000+ Real-world APIs},''
\newblock in {\em {ICLR}}, 2024.

\bibitem{shimtooldial}
Jeonghoon Shim, Gyuhyeon Seo, Cheongsu Lim, and Yohan Jo,
\newblock ``{ToolDial: Multi-turn Dialogue Generation Method for Tool-Augmented Language Models},''
\newblock in {\em {ICLR}}, 2025.

\bibitem{li2024mmedagent}
Binxu Li, Tiankai Yan, Yuanting Pan, Jie Luo, Ruiyang Ji, Jiayuan Ding, Zhe Xu, Shilong Liu, Haoyu Dong, Zihao Lin, et~al.,
\newblock ``{MMedAgent: Learning to Use Medical Tools with Multi-modal Agent},''
\newblock in {\em {EMNLP Findings}}, 2024.

\bibitem{fallahpourmedrax}
Adibvafa Fallahpour, Jun Ma, Alif Munim, Hongwei Lyu, and BO~WANG,
\newblock ``{MedRAX: Medical Reasoning Agent for Chest X-ray},''
\newblock in {\em {ICML}}, 2025.

\bibitem{heiman2025factchexcker}
Alice Heiman, Xiaoman Zhang, Emma Chen, Sung~Eun Kim, and Pranav Rajpurkar,
\newblock ``{FactCheXcker: Mitigating Measurement Hallucinations in Chest X-ray Report Generation Models},''
\newblock in {\em {CVPR}}, 2025.

\bibitem{li2023chatdoctor}
Yunxiang Li, Zihan Li, Kai Zhang, Ruilong Dan, Steve Jiang, and You Zhang,
\newblock ``{ChatDoctor: A Medical Chat Model Fine-Tuned on a Large Language Model Meta-AI (LLaMA) Using Medical Domain Knowledge},''
\newblock {\em Cureus}, vol. 15, no. 6, 2023.

\bibitem{comanici2025gemini}
Gheorghe Comanici, Eric Bieber, Mike Schaekermann, Ice Pasupat, Noveen Sachdeva, Inderjit Dhillon, Marcel Blistein, Ori Ram, Dan Zhang, Evan Rosen, et~al.,
\newblock ``{Gemini 2.5: Pushing the frontier with advanced reasoning, multimodality, long context, and next generation agentic capabilities},''
\newblock {\em arXiv:2507.06261}, 2025.

\bibitem{wagner2020ptb}
Patrick Wagner, Nils Strodthoff, Ralf-Dieter Bousseljot, Dieter Kreiseler, Fatima~I Lunze, Wojciech Samek, and Tobias Schaeffter,
\newblock ``{PTB-XL, a large publicly available electrocardiography dataset},''
\newblock {\em Scientific Data}, vol. 7, no. 1, pp. 1--15, 2020.

\bibitem{oh2022lead}
Jungwoo Oh, Hyunseung Chung, Joon-myoung Kwon, Dong-gyun Hong, and Edward Choi,
\newblock ``{Lead-agnostic self-supervised learning for local and global representations of electrocardiogram},''
\newblock in {\em {CHIL}}, 2022.

\bibitem{Makowski2021neurokit}
Dominique Makowski, Tam Pham, Zen~J. Lau, Jan~C. Brammer, Fran{\c{c}}ois Lespinasse, Hung Pham, Christopher Schölzel, and S.~H.~Annabel Chen,
\newblock ``{{NeuroKit}2: A Python toolbox for neurophysiological signal processing},''
\newblock {\em Behavior Research Methods}, vol. 53, no. 4, pp. 1689--1696, 2021.

\bibitem{chung2024time}
Hyunseung Chung, Sumin Jo, Yeonsu Kwon, and Edward Choi,
\newblock ``{Time is not enough: Time-frequency based explanation for time-series black-box models},''
\newblock in {\em {CIKM}}, 2024.

\bibitem{hu2022lora}
Edward~J Hu, Yelong Shen, Phillip Wallis, Zeyuan Allen-Zhu, Yuanzhi Li, Shean Wang, Lu~Wang, Weizhu Chen, et~al.,
\newblock ``Lora: Low-rank adaptation of large language models.,''
\newblock in {\em {ICLR}}, 2022.

\bibitem{dettmers20228bit}
Tim Dettmers, Mike Lewis, Sam Shleifer, and Luke Zettlemoyer,
\newblock ``{8-bit Optimizers via Block-wise Quantization},''
\newblock in {\em {ICLR}}, 2022.

\bibitem{unsloth}
Daniel Han, Michael Han, and {Unsloth team},
\newblock ``{Unsloth},'' 2023.

\bibitem{zheng2023judging}
Lianmin Zheng, Wei-Lin Chiang, Ying Sheng, Siyuan Zhuang, Zhanghao Wu, Yonghao Zhuang, Zi~Lin, Zhuohan Li, Dacheng Li, Eric Xing, et~al.,
\newblock ``{Judging llm-as-a-judge with mt-bench and chatbot arena},''
\newblock {\em NeurIPS}, vol. 36, 2023.

\bibitem{strodthoff2023ptb}
Nils Strodthoff, Temesgen Mehari, Claudia Nagel, Philip~J Aston, Ashish Sundar, Claus Graff, J{\o}rgen~K Kanters, Wilhelm Haverkamp, Olaf D{\"o}ssel, Axel Loewe, et~al.,
\newblock ``{PTB-XL+, a comprehensive electrocardiographic feature dataset},''
\newblock {\em Scientific Data}, vol. 10, no. 1, pp. 279, 2023.

\bibitem{liu2023g}
Yang Liu, Dan Iter, Yichong Xu, Shuohang Wang, Ruochen Xu, and Chenguang Zhu,
\newblock ``G-eval: Nlg evaluation using gpt-4 with better human alignment,''
\newblock in {\em EMNLP}, 2023, pp. 2511--2522.

\bibitem{shou2016temporal}
Zheng Shou, Dongang Wang, and Shih-Fu Chang,
\newblock ``{Temporal action localization in untrimmed videos via multi-stage cnns},''
\newblock in {\em {CVPR}}, 2016.

\end{thebibliography}
\end{document}